\pdfoutput=1
\documentclass[11pt]{article}

\usepackage[final]{acl}

\usepackage{txfonts}
\usepackage{latexsym}
\usepackage{array}
\usepackage{multirow}
\usepackage{float}
\usepackage{xcolor}
\usepackage{colortbl}
\usepackage[title]{appendix}

\usepackage[T1]{fontenc}
\usepackage[utf8]{inputenc}

\usepackage{microtype}

\usepackage{inconsolata}

\usepackage{graphicx}
\usepackage{multirow}

\title{Historical Ink: Exploring Large Language Models for Irony Detection in 19th-Century Spanish}

\author{
  Kevin Cohen$^1$ \quad Laura Manrique-Gómez$^2$ \quad Rubén Manrique$^1$  \\
  $^1$ Systems and Computing Engineering Department, Universidad de los Andes \\
  $^2$ History and Geography Department, Universidad de los Andes \\
  Bogotá D.C. \\
  \texttt{\{k.cohen, l.manriqueg, rf.manrique\}@uniandes.edu.co} \\
}

\begin{document}
\maketitle
\begin{abstract}

\end{abstract}
This study explores the use of large language models (LLMs) to enhance datasets and improve irony detection in 19th-century Latin American newspapers. Two strategies were employed to evaluate the efficacy of BERT and GPT-4o models in capturing the subtle nuances nature of irony, through both multi-class and binary classification tasks. First, we implemented dataset enhancements focused on enriching emotional and contextual cues; however, these showed limited impact on historical language analysis. The second strategy, a semi-automated annotation process, effectively addressed class imbalance and augmented the dataset with high-quality annotations. Despite the challenges posed by the complexity of irony, this work contributes to the advancement of sentiment analysis through two key contributions: introducing a new historical Spanish dataset tagged for sentiment analysis and irony detection, and proposing a semi-automated annotation methodology where human expertise is crucial for refining LLMs results, enriched by incorporating historical and cultural contexts as core features.

\section{Introduction}
Irony is a nuanced and often subtle form of communication, especially in historical texts, where cultural context is crucial in understanding the intended meaning. Detecting irony in written language has long been a challenge for natural language processing (NLP) \cite{GonzlezIbez2011}, as it is a trope whose actual meaning differs from what is
literally enunciated \cite{Hee2018}. Hence, irony detection involves identifying contradictions between what is said and the underlying meaning or context. This challenge becomes even more complex when dealing with historical texts, where linguistic expressions, cultural references, and societal norms differ significantly from modern times.

This research focuses on irony detection in 19th-century Latin American newspapers, utilizing large language models (LLMs) and feed-forward neural networks to experiment with various strategies. The dataset, annotated by experts, was processed for multi-class and binary classification. BERT-like models were used for text encoding and transfer learning, while GPT-4o models were applied for sentiment classification with customized prompts. A BERT encoder with a feed-forward neural network was used as a comparative baseline to examine the enhancement of text analysis specific to irony detection.

A semi-automatic annotation approach was also developed to incorporate new, untagged data. By employing GPT-4o models with tailored prompts, initial classifications were machine-generated and verified by human experts, reducing annotation time and effort while maintaining high-quality standards. Integrating human expertise with machine-generated results enhanced the dataset and allowed for more effective model training. Consequently, the study contributes to research on sentiment analysis in historical texts and demonstrates LLMs' capabilities in enhancing text classification in specialized contexts.

\section{Related Work}
Philosophers and linguists have yet to reach a definitive agreement on defining certain figurative tropes including irony, sarcasm, satire, hyperbole, analogy, restatement, paradox, and parody. There have been arguments regarding subtle differences, such as the humorous intention in irony versus the explicitly offensive intention in verbal sarcasm. However, there is a broader consensus on considering irony as the overarching category \cite{Kreuz2020,Colston2017}. In NLP, irony, sarcasm, and satire have often been used interchangeably, focusing on linguistic and sentiment features. Early work suggested satire exploits human tendencies like gullibility and confirmation bias \cite{Rubin2016}. Other approaches, achieving moderate success, explored linguistic patterns, including slang and profanity, using techniques such as Bi-normal Separation (BNS) \cite{Burfoot2009}. 

Traditional rule-based methods relied on linguistic indicators like interjections and hyperbole \cite{Riloff2013}. However, deep learning models such as ELMo and BERT have revolutionized the field by incorporating embeddings that capture contextual information across sentences. Rajadesingan et al. emphasized the importance of user behavior and historical tweet data in improving sarcasm classification, demonstrating how conversational context can enhance detection accuracy \cite{Rajadesingan2015}.

Both satire and sarcasm detection have significantly benefited from integrating contextual features and deep learning, showing superior performance compared to traditional methods. Nonetheless, challenges remain due to the reliance on cultural and societal context, which complicates creating highly accurate models.

Irony detection has similarly grown in importance as researchers address the challenges of figurative language. Early approaches to irony detection primarily utilized static linguistic features and manually annotated datasets but struggled to capture irony's nuanced and dynamic nature, particularly when dealing with context-specific language, such as that found in social media and historical texts.

The introduction of deep learning models using transformer-based architectures for LLMs like BERT and GPT has significantly improved the capacity for irony detection. Huang et al. explored the application of deep learning techniques such as Recurrent Neural Networks (RNN), revealing that models with attention mechanisms outperformed others on irony detection tasks using social media data \cite{Huang2017}. These models excel by capturing both syntactic and semantic relationships within sentences, enabling comprehensive irony analysis. Similarly, Ren et al. proposed a knowledge-enhanced neural network that incorporates contextual information from external knowledge sources like Wikipedia to enhance irony detection performance \cite{Ren2023}.

Recent studies have focused on improving irony detection by integrating emotional and contextual cues. Lin et al. introduced a newly developed method combining LLMs with emotion-centric text enhancement to improve the detection of irony \cite{Lin2023}. Their approach highlighted the significance of subtle emotional cues often overlooked in traditional models. By using GPT-4 to expand original texts with additional content, the researchers significantly enhanced irony detection in benchmark datasets. Ozturk et al. introduced a de-biasing approach for irony, satire, and sarcasm detection, utilizing generative LLMs to reduce stylistic biases produced by single-source corpus training datasets \cite{Ozturk2024}. Their findings indicate that such stylistic bias impacts model robustness and that LLMs-based enhancement can mitigate these biases. However, its effect on causal language models like Llama-3.1 remains limited. This approach aligns with recent trends in dataset enhancement and bias reduction to improve the detection of figurative language.

Recent methodologies introduced a human-LLMs collaborative annotation framework that addresses the limitations of LLMs-generated labels by combining automated annotation with human expertise \cite{Wang2024}. Their method incorporates three major steps: LLMs predict labels and generate explanations; a verifier assesses label quality; and human annotators review and re-annotate labels flagged as low quality. This research shows the necessity for hybrid approaches that merge LLMs scalability with human precision, particularly for complex tasks such as irony detection. 

Despite advancements, challenges persist in handling class imbalance, stylistic bias, and the nuanced nature of figurative language, especially in historical texts. Prior work on deep learning models and LLMs-driven enhancements often focuses on contemporary datasets, neglecting historical linguistic and cultural context variations. Moreover, research on semi-automated annotation strategies leveraging human-LLMs collaboration is limited. Our study addresses these gaps by integrating a structured semi-automated annotation process, improving dataset balance, and fine-tuning domain-specific models for 19th-century Spanish. By combining LLMs-powered augmentation with human verification, we propose a scalable method for enriching training data while maintaining historical linguistic authenticity. This work refines irony detection in historical texts and provides a broader framework for augmenting and improving figurative language classification in underrepresented domains.

\section{Data}
The dataset used in this research initially comprised a tagged corpus of 2,734 entries. Each entry contains text samples randomly extracted from 19th-century Latin American newspapers— the LatamXIX Dataset \cite{Manrique_Gomez2024}. Tags were manually assigned by three human experts to each entry, falling into one of the following categories: "IRONÍA" (irony), "POSITIVO" (positive), "NEGATIVO" (negative), and "NEUTRO" (neutral). These tags represent the predominant sentiment of the extracted text. In our dataset, irony involves multilayered expressions of emotions, including criticism, humor, and sarcasm, as well as the use of poetic language —a distinctive feature of the era. 

The primary dataset was used to conduct experiments on text enhancement and fine-tune BERT-like models for classifying irony. As mentioned previously, the dataset comprises four distinct classes. To extend the scope of experimentation, a copy of the dataset was created where the "POSITIVO," "NEGATIVO," and "NEUTRO" classes were merged into a single class labeled "NO IRONÍA" (not irony). This transformation converts the task from multi-class classification to binary classification. For better computational processing, a new column named "category\_encoded" was created in both datasets to contain the same tags encoded as numerical values, facilitating interpretation.

In the second phase of the research, the primary dataset with the original text (without enhancement) was augmented with 1,016 additional entries. These new fragments also originated from the LatamXIX Dataset and were used to implement the semi-automated methodology for annotation. 

The final dataset consists of 3,750 annotated entries, resulting in a more balanced collection that corrected the initial underrepresentation of the "IRONÍA" class, preserves the historical linguistic value of the original texts, and improves the accuracy of the LLMs BERT-like models fine-tuned for the historical irony classification task\footnote{The dataset is available at \url{https://huggingface.co/datasets/Flaglab/latam-xix-tagged-augmented} in its three versions: "primary", "enhanced", and "augmented"}.

\section{Methodology}

The experimentation encompasses two primary aspects: dataset enhancement and augmentation and the construction of the classification pipeline. GPT-4o was employed in conjunction with prompt engineering to enhance the text and establish a baseline for measuring improvements in the classification task. Subsequently, BERT-like models were used to classify the dataset. Detailed explanations of these components are provided in the following subsections.

\subsection{Dataset Enhancing}
A key objective of this research was to evaluate how models like GPT-4o can enhance context and text to improve sentiment analysis, specifically focusing on detecting historical irony. Several prompts were developed and tested to achieve satisfactory results. The evaluation involved a small, balanced dataset of 40 diverse entries from the original dataset. Each prompt's performance was individually analyzed to ensure that the responses were closely aligned with the manual classification of the original data. A prompt deemed effective in performing the task was then applied to the entire dataset alongside a neural network.

The final prompt used to enhance the dataset was:
\textit{"Expand this text while preserving its original meaning, placing a strong emphasis on its emotional content to enhance the identification of its overall sentiment. Respond only with the expanded text, and strive to maintain the syntax and morphology characteristic of 19th-century Latin American Spanish." }    

The prompt, originally in Spanish as detailed in Appendix \ref{prompt1}, does not reference irony or specific sentiments to avoid bias that pre-classified data might introduce. This approach allows for an initial observation of how GPT-4o expands the original texts. Appendix \ref{propmtex1} provides an example of the  GPT-4o input and output text generated from the data enhancement process.

\subsection{Dataset Augmentation }
In addition to enhancing the original dataset, a new strategy was introduced to include previously untagged data with a high potential for irony detection. This involved designing a prompt for processing 1,034 new entries, selected from sources likely to contain ironic content.

The dataset was classified using GPT-4o to identify scenarios involving "IRONÍA," "POSITIVO," "NEGATIVO," and "NEUTRO." Appendix \ref{ex1} details the prompt used for this task, and Appendix \ref{propmtex}  provides an example of the  GPT-4o output tag and explanation generated in the augmentation process.
    
The prompt design featured four key components: [1] Context: Oriented the model to analyze 19th-century Spanish texts from Latin American newspapers. [2] Irony Recognition: Guided the model to recognize contradictions in three scenarios:
        \begin{itemize}
        \item Between described reality and expression.
        \item Historical reality and expression.
        \item Expression tone as indicated by capitalization and punctuation.
    \end{itemize} 

[3] Exceptions: Addressed frequent misclassifications by instructing the model not to mark irony in cases of: 
        \begin{itemize}
        \item Political opinions.
        \item Poetic language.
        \item Instances lacking humorous intent.
    \end{itemize} 
        
Finally, [4] the Task: Required explanations for classification decisions to be appended in asterisks (*), facilitating sentiment extraction via regular expressions.
    
Each entry, processed by GPT-4o using this prompt, produced an output with an assigned tag (e.g., 'IRONÍA') and an explanation for the classification (e.g., The text contains contradictory statements suggesting irony). An expert reviewed these outputs to verify accuracy, significantly reducing the time and effort compared to manual annotation from scratch.

\begin{figure*}\centering
    \includegraphics[width=0.8\linewidth]{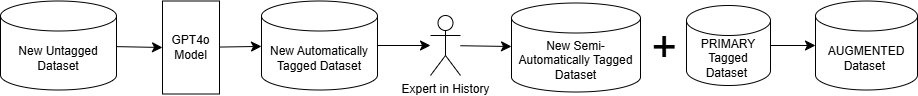}
    \caption{Semi-Automatic Annotation Methodology.}
    \label{fig:DA}
\end{figure*}

The human verification process identified 18 entries with low-quality OCR transcription. These 'unreadable' entries were excluded from the final dataset augmentation. The verified new sample resulted in 1,016 entries added to the primary dataset. The augmented dataset was then used to fine-tune the model, detailed in the next section, to enhance its performance in irony detection tasks. Figure \ref{fig:DA} illustrates this semi-automatic annotation methodology, and Table \ref{tab:my-table60} summarizes the datasets mentioned:

\begin{table}[H]
\resizebox{\columnwidth}{!}{%
\begin{tabular}{|c|c|c|c|cll|}
\hline
\textbf{Dataset} & \textbf{Num. Entries} & \textbf{Augmented} & \textbf{Enhanced} & \multicolumn{3}{c|}{\textbf{Experiment}}       \\ \hline
PRIMARY              & 2734                       & NO                 & NO                & \multicolumn{3}{c|}{Baseline}                  \\ \hline
ENHANCED             & 2734                       & NO                 & \textbf{YES}      & \multicolumn{3}{c|}{Prompt Based Enhancement}  \\ \hline
AUGMENTED             & 3750                       & \textbf{YES}       & NO                & \multicolumn{3}{c|}{Semi-Automatic Annotation} \\ \hline
\end{tabular}%
}
\caption{Datasets Summary. \textit{PRIMARY:} Primary Human Annotated Dataset. \textit{ENHANCED:} Enhanced Dataset Automatic Annotated. \textit{AUGMENTED:} Augmented Semi-automatic Annotated Dataset.}
\label{tab:my-table60}
\end{table}

\subsection{BERT-based Classification Pipeline}
The architecture designed for irony classification consists of an LLMs BERT encoder with a feedforward neural network head. The model comprises three layers with the following features:  

\begin{itemize}
    \item An input layer of size 768, selected to match the standard dimensions of the contextual vector representations produced by BERT-like models.
    \item A first hidden layer employing a ReLU activation function, with a weight matrix of dimensions 768 x 50.
    \item A fully connected layer that maps the hidden layer to the output layer, with a weight matrix of dimensions 50 x output\_dim, where output\_dim can be either four or two, depending on whether the classification is multi-class or binary.
\end{itemize}

\begin{figure*}\centering
    \includegraphics[width=0.8\linewidth]{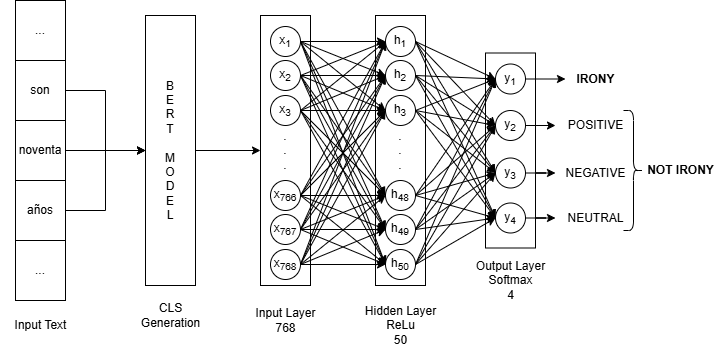}
    \caption{Architecture of the BERT-based Classification Pipeline.}
    \label{fig:architecture}
\end{figure*}

As shown in Figure \ref{fig:architecture}, the architecture uses the contextual embedding generated by a BERT family model for text representation. At the bottom of each layer, the corresponding dimensions and activation functions are indicated. In the model's output layer, the dimensions vary based on the type of classification (binary or multi-class). The activation function is a sigmoid for binary classification, and the number of nodes is adjusted to two. The following BERT-like models were evaluated:

    \begin{itemize}
        \item bert-base-uncased
        \item bert-base-multilingual-uncased
        \item dccuchile/bert-base-spanish-wwm-uncased
        \item dccuchile/bert-base-spanish-wwm-cased
        \item beto-cased-finetuned-xix-latam. 
    \end{itemize}

The selection included standard BERT models, both base and multilingual versions, as well as models tailored for contemporary Spanish, and a version trained on 19th-century Spanish texts \cite{montes2024}.

The training process was configured to run for a maximum of 1500 epochs. However, in practice, training was automatically halted when the validation loss began to diverge significantly from the training loss. This mechanism was employed to prevent overfitting.

\subsection{Experiments}
The experiments were evaluated in three main phases, each designed to assess different strategies for irony detection. The first phase establishes \textbf{baselines} using the \textit{PRIMARY} dataset (see Table \ref{tab:my-table60}). This phase consists of two separate evaluations: first, GPT-4o is used to directly classify the samples based on a prompt that provides context, defines irony, outlines exceptions, and specifies the task, as detailed in Appendix \ref{ex1}. The initial prompt-based tagging made by GPT-4o was followed by independent processing with the BERT-based Classification Pipeline (see Figure \ref{fig:architecture}). These baselines served as reference points to compare the impact of dataset enhancements and augmentations.

In the second phase, we tested the hypothesis that GPT-4o can \textbf{enhance} the original text to facilitate classification. In this experiment, GPT-4o enriched the emotional and contextual features of the historical texts, generating the \textit{ENHANCED} dataset (see Table \ref{tab:my-table60}). This dataset was then processed using the BERT-based Classification Pipeline to assess whether the additional contextual and emotional cues improved classification performance.

The final phase involved \textbf{augmenting} the dataset through a semi-automated annotation process. GPT-4o classified entries in the new samples, providing labels and detailed justifications. Human experts reviewed these automatic annotations to ensure accuracy and preserve the historical value of the dataset. The newly verified entries were integrated into the PRIMARY dataset, generating the \textit{AUGMENTED} dataset (see Table \ref{tab:my-table60}), which was subsequently processed using the BERT-based Classification Pipeline. In this final phase, only the top three performing BERT-like models from the earlier experiments were used for classification.

After each phase, the neural network results were evaluated using precision, recall, accuracy, and F1 score metrics. Testing was conducted separately for both binary and multi-class classifications. 

\section{Results}
\subsection{Baselines}

The results obtained using GPT-4o with the PRIMARY dataset are presented in Table \ref{tab:my-table1}.

\begin{table}[H]
\resizebox{\columnwidth}{!}{%
\begin{tabular}{|c|l|c|c|c|c|}
\hline
\textit{\textbf{Model}}                                                & \multicolumn{1}{c|}{\textit{\textbf{Category}}} & \textit{\textbf{Precision}} & \textit{\textbf{Recall}} & \textit{\textbf{F1 Score}} & \textit{\textbf{Accuracy}}                     \\ \hline
\rowcolor[HTML]{FFFFFF} 
\cellcolor[HTML]{FFFFFF}                                               & IRONY                                           & 0.24                        & 0.80                     & 0.37                       & \cellcolor[HTML]{FFFFFF}                       \\ \cline{2-5}
\rowcolor[HTML]{FFFFFF} 
\cellcolor[HTML]{FFFFFF}                                               & NEGATIVE                                        & 0.55                        & 0.33                     & 0.41                       & \cellcolor[HTML]{FFFFFF}                       \\ \cline{2-5}
\rowcolor[HTML]{FFFFFF} 
\cellcolor[HTML]{FFFFFF}                                               & NEUTRAL                                         & 0.44                        & 0.76                     & 0.55                       & \cellcolor[HTML]{FFFFFF}                       \\ \cline{2-5}
\rowcolor[HTML]{FFFFFF} 
\cellcolor[HTML]{FFFFFF}                                               & POSITIVE                                        & 0.86                        & 0.01                     & 0.03                       & \cellcolor[HTML]{FFFFFF}                       \\ \cline{2-5}
\rowcolor[HTML]{FFFFFF} 
\multirow{-5}{*}{\cellcolor[HTML]{FFFFFF}\textbf{Base GPT 4o -Prompt}} & W. AVG                                          & 0.60                        & 0.39                     & 0.31                       & \multirow{-5}{*}{\cellcolor[HTML]{FFFFFF}0.39} \\ \hline
\end{tabular}%
}
\caption{\textit{Results for the multi-class classification task using GPT-4o with the prompt specified in Appendix \ref{ex1}.}}
\label{tab:my-table1}
\end{table}

The results indicate that the 'IRONY' class achieved a precision of only 0.24, with an overall accuracy of 0.39. Thus, relying solely on GPT-4o models and prompting is not a viable approach for classifying historical Spanish texts.

\begin{table}[H]
\resizebox{\columnwidth}{!}{%
\begin{tabular}{|c|l|c|c|c|c|}
\hline
\textit{\textbf{Model}}                                                & \multicolumn{1}{c|}{\textit{\textbf{Category}}}     & \textit{\textbf{Precision}}                    & \textit{\textbf{Recall}}                       & \textit{\textbf{F1 Score}}                     & \textit{\textbf{Accuracy}}                     \\ \hline
\rowcolor[HTML]{FFFFFF} 
\cellcolor[HTML]{FFFFFF}                                               & \cellcolor[HTML]{FFFFFF}                            & \cellcolor[HTML]{FFFFFF}                       & \cellcolor[HTML]{FFFFFF}                       & \cellcolor[HTML]{FFFFFF}                       & \cellcolor[HTML]{FFFFFF}                       \\
\rowcolor[HTML]{FFFFFF} 
\cellcolor[HTML]{FFFFFF}                                               & \multirow{-2}{*}{\cellcolor[HTML]{FFFFFF}IRONY}     & \multirow{-2}{*}{\cellcolor[HTML]{FFFFFF}0.24} & \multirow{-2}{*}{\cellcolor[HTML]{FFFFFF}0.80} & \multirow{-2}{*}{\cellcolor[HTML]{FFFFFF}0.37} & \cellcolor[HTML]{FFFFFF}                       \\ \cline{2-5}
\rowcolor[HTML]{FFFFFF} 
\cellcolor[HTML]{FFFFFF}                                               & \cellcolor[HTML]{FFFFFF}                            & \cellcolor[HTML]{FFFFFF}                       & \cellcolor[HTML]{FFFFFF}                       & \cellcolor[HTML]{FFFFFF}                       & \cellcolor[HTML]{FFFFFF}                       \\
\rowcolor[HTML]{FFFFFF} 
\cellcolor[HTML]{FFFFFF}                                               & \multirow{-2}{*}{\cellcolor[HTML]{FFFFFF}NOT IRONY} & \multirow{-2}{*}{\cellcolor[HTML]{FFFFFF}0.97} & \multirow{-2}{*}{\cellcolor[HTML]{FFFFFF}0.71} & \multirow{-2}{*}{\cellcolor[HTML]{FFFFFF}0.82} & \cellcolor[HTML]{FFFFFF}                       \\ \cline{2-5}
\rowcolor[HTML]{FFFFFF} 
\multirow{-5}{*}{\cellcolor[HTML]{FFFFFF}\textbf{Base GPT 4o -Prompt}} & AVG                                                 & 0.60                                           & 0.75                                           & 0.59                                           & \multirow{-5}{*}{\cellcolor[HTML]{FFFFFF}0.72} \\ \hline
\end{tabular}%
}
\caption{\textit{Results for binary classification task using GPT-4o with the prompt specified in Appendix \ref{ex1}.}}
\label{tab:my-table2}
\end{table}

The' NOT IRONY' class showed improvement in the binary classification scenario (see Table \ref{tab:my-table2}). Although the model performed better when detecting non-ironic situations.

Next, we discuss the results obtained using the complete BERT-based Classification Pipeline described in the methodology, which includes contextual embeddings from BERT-family encoder models. The best-performing encoder results are presented in Table \ref{tab:my-table3}, with a comprehensive list of all tested encoders in Appendix \ref{ex5}.

\begin{table}[H]
\resizebox{\columnwidth}{!}{%
\begin{tabular}{|c|l|c|c|c|c|}
\hline
\textit{\textbf{Model}}                                                                    & \multicolumn{1}{c|}{\textit{\textbf{Category}}} & \textit{\textbf{Precision}} & \textit{\textbf{Recall}} & \textit{\textbf{F1 Score}} & \textit{\textbf{Accuracy}}                     \\ \hline

\rowcolor[HTML]{FFFFFF} 
\cellcolor[HTML]{FFFFFF}                                                                   & IRONY                                           & 0.61                        & 0.47                     & 0.53                       & \cellcolor[HTML]{FFFFFF}                       \\ \cline{2-5}
\rowcolor[HTML]{FFFFFF} 
\cellcolor[HTML]{FFFFFF}                                                                   & NEGATIVE                                        & 0.60                        & 0.62                     & 0.61                       & \cellcolor[HTML]{FFFFFF}                       \\ \cline{2-5}
\rowcolor[HTML]{FFFFFF} 
\cellcolor[HTML]{FFFFFF}                                                                   & NEUTRAL                                         & 0.72                        & 0.66                     & 0.69                       & \cellcolor[HTML]{FFFFFF}                       \\ \cline{2-5}
\rowcolor[HTML]{FFFFFF} 
\cellcolor[HTML]{FFFFFF}                                                                   & POSITIVE                                        & 0.66                        & 0.75                     & 0.70                       & \cellcolor[HTML]{FFFFFF}                       \\ \cline{2-5}
\rowcolor[HTML]{FFFFFF} 
\multirow{-5}{*}{\cellcolor[HTML]{FFFFFF}\textbf{dccuchile/bert-base-spanish-wwm-cased}}   & W. AVG                                          & 0.66                        & 0.66                     & 0.65                       & \multirow{-5}{*}{\cellcolor[HTML]{FFFFFF}0.66} \\ \hline

\end{tabular}%
}
\caption{\textit{Results of the BERT-based Classification Pipeline. The table presents only the best-performing encoder model for the multi-class task.}}
\label{tab:my-table3}
\end{table}

Table \ref{tab:my-table3} shows an accuracy of 0.66, effectively detecting the 'IRONY' class and other sentiment categories. Although not extraordinary, these results show notable improvement over the GPT-4o classification outcomes. The results in the binary classification scenario, as shown in Table \ref{tab:my-table4}, corroborate this trend. The model achieved near-perfect classification for the 'NOT IRONY' category. However, these results should be interpreted cautiously due to the potential bias from the underrepresentation of the "IRONY" class in the PRIMARY dataset. Additionally, significant room for improvement in classifying the 'IRONY' category remains. These results, obtained using the neural network with the "dccuchile/bert-base-spanish-wwm-case" encoder, serve as the benchmark for subsequent experiments.

\begin{table}[H]
\resizebox{\columnwidth}{!}{%
\begin{tabular}{|c|l|c|c|c|c|}
\hline
\textit{\textbf{Model}}                                                                    & \multicolumn{1}{c|}{\textit{\textbf{Category}}}     & \textit{\textbf{Precision}}                    & \textit{\textbf{Recall}}                       & \textit{\textbf{F1 Score}}                     & \textit{\textbf{Accuracy}}                     \\ \hline

\rowcolor[HTML]{FFFFFF} 
\cellcolor[HTML]{FFFFFF}                                                                   & \cellcolor[HTML]{FFFFFF}                            & \cellcolor[HTML]{FFFFFF}                       & \cellcolor[HTML]{FFFFFF}                       & \cellcolor[HTML]{FFFFFF}                       & \cellcolor[HTML]{FFFFFF}                       \\
\rowcolor[HTML]{FFFFFF} 
\cellcolor[HTML]{FFFFFF}                                                                   & \multirow{-2}{*}{\cellcolor[HTML]{FFFFFF}IRONY}     & \multirow{-2}{*}{\cellcolor[HTML]{FFFFFF}0.80} & \multirow{-2}{*}{\cellcolor[HTML]{FFFFFF}0.34} & \multirow{-2}{*}{\cellcolor[HTML]{FFFFFF}0.48} & \cellcolor[HTML]{FFFFFF}                       \\ \cline{2-5}
\rowcolor[HTML]{FFFFFF} 
\cellcolor[HTML]{FFFFFF}                                                                   & \cellcolor[HTML]{FFFFFF}                            & \cellcolor[HTML]{FFFFFF}                       & \cellcolor[HTML]{FFFFFF}                       & \cellcolor[HTML]{FFFFFF}                       & \cellcolor[HTML]{FFFFFF}                       \\
\rowcolor[HTML]{FFFFFF} 
\cellcolor[HTML]{FFFFFF}                                                                   & \multirow{-2}{*}{\cellcolor[HTML]{FFFFFF}NOT IRONY} & \multirow{-2}{*}{\cellcolor[HTML]{FFFFFF}0.92} & \multirow{-2}{*}{\cellcolor[HTML]{FFFFFF}0.99} & \multirow{-2}{*}{\cellcolor[HTML]{FFFFFF}0.95} & \cellcolor[HTML]{FFFFFF}                       \\ \cline{2-5}
\rowcolor[HTML]{FFFFFF} 
\multirow{-5}{*}{\cellcolor[HTML]{FFFFFF}\textbf{dccuchile/bert-base-spanish-wwm-cased}}   & AVG                                                 & 0.86                                           & 0.66                                           & 0.72                                           & \multirow{-5}{*}{\cellcolor[HTML]{FFFFFF}0.91} \\ \hline

\end{tabular}
}
\caption{\textit{Results of the BERT-based Classification Pipeline. The table presents only the best-performing encoder model for the binary task.}}
\label{tab:my-table4}
\end{table}

\subsection{Enhancement}

This section presents the results obtained using the \textit{ENHANCED} dataset and the BERT-based classification pipeline. Table \ref{tab:my-table5} reports the results for the multiclass scenario, while Table \ref{tab:my-table6} shows the binary classification results for the best-performing encoder. Appendix \ref{ex6} shows a complete set of tabulated results.

\begin{table}[H]
\resizebox{\columnwidth}{!}{%
\begin{tabular}{|
>{\columncolor[HTML]{FFFFFF}}c |
>{\columncolor[HTML]{FFFFFF}}l |
>{\columncolor[HTML]{FFFFFF}}c |
>{\columncolor[HTML]{FFFFFF}}c |
>{\columncolor[HTML]{FFFFFF}}c |
>{\columncolor[HTML]{FFFFFF}}c |}
\hline
\textit{\textbf{Model}}                                                                    & \multicolumn{1}{c|}{\cellcolor[HTML]{FFFFFF}\textit{\textbf{Category}}} & \textit{Precision} & \textit{Recall} & \textit{F1 Score} & \textit{Accuracy}                              \\ \hline
\cellcolor[HTML]{FFFFFF}                                                                   & IRONY                                                                   & \textbf{0.65}      & \textbf{0.51}   & \textbf{0.57}     & \cellcolor[HTML]{FFFFFF}                       \\ \cline{2-5}
\cellcolor[HTML]{FFFFFF}                                                                   & NEGATIVE                                                                & 0.54               & \textbf{0.62}   & \textbf{0.58}     & \cellcolor[HTML]{FFFFFF}                       \\ \cline{2-5}
\cellcolor[HTML]{FFFFFF}                                                                   & NEUTRAL                                                                 & 0.69               & 0.53            & 0.60              & \cellcolor[HTML]{FFFFFF}                       \\ \cline{2-5}
\cellcolor[HTML]{FFFFFF}                                                                   & POSITIVE                                                                & 0.58               & 0.69            & 0.63              & \cellcolor[HTML]{FFFFFF}                       \\ \cline{2-5}
\multirow{-5}{*}{\cellcolor[HTML]{FFFFFF}\textbf{beto-cased-finetuned-xix-latam}}                      & W. AVG                                                                  & 0.61               & 0.60            & 0.60              & \multirow{-5}{*}{\cellcolor[HTML]{FFFFFF}0.60} \\ \hline
\end{tabular}%
}
\caption{\textit{Results of the BERT-based Classification Pipeline on the \textit{ENHANCED} dataset. The table presents only the best-performing encoder model for the multi-class task.}}
\label{tab:my-table5}
\end{table}

In the multiclass scenario, while the 'IRONY' class slightly improved over the baseline, the overall performance remained similar. The 'IRONY' classification exhibited greater reliability in binary classification, without a clear improvement over the baseline. These results may be influenced by the unique characteristics of historical texts and the challenges the GPT-4o model faces in capturing deeper emotional and contextual cues.

\begin{table}[H]
\resizebox{\columnwidth}{!}{%
\begin{tabular}{|
>{\columncolor[HTML]{FFFFFF}}c |
>{\columncolor[HTML]{FFFFFF}}l |
>{\columncolor[HTML]{FFFFFF}}c |
>{\columncolor[HTML]{FFFFFF}}c |
>{\columncolor[HTML]{FFFFFF}}c |
>{\columncolor[HTML]{FFFFFF}}c |}
\hline
\textit{\textbf{Model}}                                                                    & \multicolumn{1}{c|}{\cellcolor[HTML]{FFFFFF}\textit{\textbf{Category}}} & \textit{\textbf{Precision}}                             & \textit{\textbf{Recall}}                                & \textit{\textbf{F1 Score}}                              & \textit{\textbf{Accuracy}}                              \\ \hline

\cellcolor[HTML]{FFFFFF}                                                                   & \cellcolor[HTML]{FFFFFF}                                                & \cellcolor[HTML]{FFFFFF}                                & \cellcolor[HTML]{FFFFFF}                                & \cellcolor[HTML]{FFFFFF}                                & \cellcolor[HTML]{FFFFFF}                                \\
\cellcolor[HTML]{FFFFFF}                                                                   & \multirow{-2}{*}{\cellcolor[HTML]{FFFFFF}IRONY}                         & \multirow{-2}{*}{\cellcolor[HTML]{FFFFFF}0.65}          & \multirow{-2}{*}{\cellcolor[HTML]{FFFFFF}\textbf{0.32}} & \multirow{-2}{*}{\cellcolor[HTML]{FFFFFF}\textbf{0.43}} & \cellcolor[HTML]{FFFFFF}                                \\ \cline{2-5}
\cellcolor[HTML]{FFFFFF}                                                                   & \cellcolor[HTML]{FFFFFF}                                                & \cellcolor[HTML]{FFFFFF}                                & \cellcolor[HTML]{FFFFFF}                                & \cellcolor[HTML]{FFFFFF}                                & \cellcolor[HTML]{FFFFFF}                                \\
\cellcolor[HTML]{FFFFFF}                                                                   & \multirow{-2}{*}{\cellcolor[HTML]{FFFFFF}NOT IRONY}                     & \multirow{-2}{*}{\cellcolor[HTML]{FFFFFF}0.92}          & \multirow{-2}{*}{\cellcolor[HTML]{FFFFFF}0.98}          & \multirow{-2}{*}{\cellcolor[HTML]{FFFFFF}0.95}          & \cellcolor[HTML]{FFFFFF}                                \\ \cline{2-5}
\multirow{-5}{*}{\cellcolor[HTML]{FFFFFF}\textbf{dccuchile/bert-base-spanish-wwm-uncased}} & AVG                                                                     & 0.78                                                    & \textbf{0.65}                                           & \textbf{0.69}                                           & \multirow{-5}{*}{\cellcolor[HTML]{FFFFFF}0.90}          \\ \hline

\end{tabular}%
}
\caption{\textit{Results of the BERT-based Classification Pipeline on the \textit{ENHANCED} dataset. The table presents only the best-performing encoder model for the binary task.}}
\label{tab:my-table6}
\end{table}

\subsection{Augmentation}
The following tables present the results obtained when experimenting with the \textit{AUGMENTED} dataset, using semi-automatically annotated data. As mentioned previously, only the top three encoder models, according to 'IRONY' class metrics, were considered in this experiment.

\begin{table}[H]
\resizebox{\columnwidth}{!}{%
\begin{tabular}{|c|l|c|c|c|c|}
\hline
\textit{\textbf{Model}}                                                                    & \multicolumn{1}{c|}{\textit{\textbf{Category}}} & \textit{\textbf{Precision}} & \textit{\textbf{Recall}} & \textit{\textbf{F1 Score}} & \textit{\textbf{Accuracy}}                     \\ \hline
\rowcolor[HTML]{FFFFFF} 
\cellcolor[HTML]{FFFFFF}                                                                   & IRONY                                           & \textbf{0.65}               & \textbf{0.72}            & \textbf{0.68}              & \cellcolor[HTML]{FFFFFF}                       \\ \cline{2-5}
\rowcolor[HTML]{FFFFFF} 
\cellcolor[HTML]{FFFFFF}                                                                   & NEGATIVE                                        & 0.53                        & \textbf{0.59}            & 0.56                       & \cellcolor[HTML]{FFFFFF}                       \\ \cline{2-5}
\rowcolor[HTML]{FFFFFF} 
\cellcolor[HTML]{FFFFFF}                                                                   & NEUTRAL                                         & 0.68                        & \textbf{0.63}            & 0.65                       & \cellcolor[HTML]{FFFFFF}                       \\ \cline{2-5}
\rowcolor[HTML]{FFFFFF} 
\cellcolor[HTML]{FFFFFF}                                                                   & POSITIVE                                        & \textbf{0.66}               & 0.61                     & 0.64                       & \cellcolor[HTML]{FFFFFF}                       \\ \cline{2-5}
\rowcolor[HTML]{FFFFFF} 
\multirow{-5}{*}{\cellcolor[HTML]{FFFFFF}\textbf{dccuchile/bert-base-spanish-wwm-uncased}} & W. AVG                                          & \textbf{0.64}               & 0.63                     & 0.63                       & \multirow{-5}{*}{\cellcolor[HTML]{FFFFFF}0.63} \\ \hline
\rowcolor[HTML]{FFFFFF} 
\cellcolor[HTML]{FFFFFF}                                                                   & IRONY                                           & \textbf{0.65}               & \textbf{0.67}            & \textbf{0.66}              & \cellcolor[HTML]{FFFFFF}                       \\ \cline{2-5}
\rowcolor[HTML]{FFFFFF} 
\cellcolor[HTML]{FFFFFF}                                                                   & NEGATIVE                                        & 0.51                        & 0.56                     & 0.53                       & \cellcolor[HTML]{FFFFFF}                       \\ \cline{2-5}
\rowcolor[HTML]{FFFFFF} 
\cellcolor[HTML]{FFFFFF}                                                                   & NEUTRAL                                         & 0.67                        & 0.61                     & 0.64                       & \cellcolor[HTML]{FFFFFF}                       \\ \cline{2-5}
\rowcolor[HTML]{FFFFFF} 
\cellcolor[HTML]{FFFFFF}                                                                   & POSITIVE                                        & 0.62                        & 0.61                     & 0.62                       & \cellcolor[HTML]{FFFFFF}                       \\ \cline{2-5}
\rowcolor[HTML]{FFFFFF} 
\multirow{-5}{*}{\cellcolor[HTML]{FFFFFF}\textbf{dccuchile/bert-base-spanish-wwm-cased}}   & W. AVG                                          & 0.59                        & 0.59                     & 0.59                       & \multirow{-5}{*}{\cellcolor[HTML]{FFFFFF}0.61} \\ \hline
\rowcolor[HTML]{FFFFFF} 
\cellcolor[HTML]{FFFFFF}                                                                   & IRONY                                           & \textbf{0.70}               & \textbf{0.66}            & \textbf{0.68}              & \cellcolor[HTML]{FFFFFF}                       \\ \cline{2-5}
\rowcolor[HTML]{FFFFFF} 
\cellcolor[HTML]{FFFFFF}                                                                   & NEGATIVE                                        & 0.50                        & 0.55                     & 0.52                       & \cellcolor[HTML]{FFFFFF}                       \\ \cline{2-5}
\rowcolor[HTML]{FFFFFF} 
\cellcolor[HTML]{FFFFFF}                                                                   & NEUTRAL                                         & 0.64                        & \textbf{0.64}            & \textbf{0.64}              & \cellcolor[HTML]{FFFFFF}                       \\ \cline{2-5}
\rowcolor[HTML]{FFFFFF} 
\cellcolor[HTML]{FFFFFF}                                                                   & POSITIVE                                        & \textbf{0.62}               & 0.60                     & 0.61                       & \cellcolor[HTML]{FFFFFF}                       \\ \cline{2-5}
\rowcolor[HTML]{FFFFFF} 
\multirow{-5}{*}{\cellcolor[HTML]{FFFFFF}\textbf{beto-cased-finetuned-xix-latam}}          & W. AVG                                          & 0.62                        & 0.61                     & 0.62                       & \multirow{-5}{*}{\cellcolor[HTML]{FFFFFF}0.61} \\ \hline
\end{tabular}%
}
\caption{\textit{Results of the BERT-based Classification Pipeline on the \textit{AUGMENTED} dataset for the multiclass task.}}
\label{tab:my-table7}
\end{table}

Unlike in previous experiments, the results presented in Table \ref{tab:my-table7} show more promising improvements for the IRONY class. While the precision metric was similar to that from Table \ref{tab:my-table6}, recall was significantly improved. This indicates that with the training data, the model can detect irony in more cases without compromising precision.

\begin{table}[H]
\resizebox{\columnwidth}{!}{%
\begin{tabular}{|c|l|c|c|c|c|}
\hline
\textit{\textbf{Model}}                                                                    & \multicolumn{1}{c|}{\textit{\textbf{Category}}}     & \textit{\textbf{Precision}}                             & \textit{\textbf{Recall}}                                & \textit{\textbf{F1 Score}}                              & \textit{\textbf{Accuracy}}                     \\ \hline
\rowcolor[HTML]{FFFFFF} 
\cellcolor[HTML]{FFFFFF}                                                                   & \cellcolor[HTML]{FFFFFF}                            & \cellcolor[HTML]{FFFFFF}                                & \cellcolor[HTML]{FFFFFF}                                & \cellcolor[HTML]{FFFFFF}                                & \cellcolor[HTML]{FFFFFF}                       \\
\rowcolor[HTML]{FFFFFF} 
\cellcolor[HTML]{FFFFFF}                                                                   & \multirow{-2}{*}{\cellcolor[HTML]{FFFFFF}IRONY}     & \multirow{-2}{*}{\cellcolor[HTML]{FFFFFF}0.65}          & \multirow{-2}{*}{\cellcolor[HTML]{FFFFFF}\textbf{0.62}} & \multirow{-2}{*}{\cellcolor[HTML]{FFFFFF}\textbf{0.64}} & \cellcolor[HTML]{FFFFFF}                       \\ \cline{2-5}
\rowcolor[HTML]{FFFFFF} 
\cellcolor[HTML]{FFFFFF}                                                                   & \cellcolor[HTML]{FFFFFF}                            & \cellcolor[HTML]{FFFFFF}                                & \cellcolor[HTML]{FFFFFF}                                & \cellcolor[HTML]{FFFFFF}                                & \cellcolor[HTML]{FFFFFF}                       \\
\rowcolor[HTML]{FFFFFF} 
\cellcolor[HTML]{FFFFFF}                                                                   & \multirow{-2}{*}{\cellcolor[HTML]{FFFFFF}NOT IRONY} & \multirow{-2}{*}{\cellcolor[HTML]{FFFFFF}0.90}          & \multirow{-2}{*}{\cellcolor[HTML]{FFFFFF}0.91}          & \multirow{-2}{*}{\cellcolor[HTML]{FFFFFF}0.91}          & \cellcolor[HTML]{FFFFFF}                       \\ \cline{2-5}
\rowcolor[HTML]{FFFFFF} 
\multirow{-5}{*}{\cellcolor[HTML]{FFFFFF}\textbf{dccuchile/bert-base-spanish-wwm-uncased}} & AVG                                                 & 0.78                                                    & \textbf{0.77}                                           & \textbf{0.77}                                           & \multirow{-5}{*}{\cellcolor[HTML]{FFFFFF}0.85} \\ \hline
\rowcolor[HTML]{FFFFFF} 
\cellcolor[HTML]{FFFFFF}                                                                   & \cellcolor[HTML]{FFFFFF}                            & \cellcolor[HTML]{FFFFFF}                                & \cellcolor[HTML]{FFFFFF}                                & \cellcolor[HTML]{FFFFFF}                                & \cellcolor[HTML]{FFFFFF}                       \\
\rowcolor[HTML]{FFFFFF} 
\cellcolor[HTML]{FFFFFF}                                                                   & \multirow{-2}{*}{\cellcolor[HTML]{FFFFFF}IRONY}     & \multirow{-2}{*}{\cellcolor[HTML]{FFFFFF}0.72}          & \multirow{-2}{*}{\cellcolor[HTML]{FFFFFF}\textbf{0.70}} & \multirow{-2}{*}{\cellcolor[HTML]{FFFFFF}\textbf{0.71}} & \cellcolor[HTML]{FFFFFF}                       \\ \cline{2-5}
\rowcolor[HTML]{FFFFFF} 
\cellcolor[HTML]{FFFFFF}                                                                   & \cellcolor[HTML]{FFFFFF}                            & \cellcolor[HTML]{FFFFFF}                                & \cellcolor[HTML]{FFFFFF}                                & \cellcolor[HTML]{FFFFFF}                                & \cellcolor[HTML]{FFFFFF}                       \\
\rowcolor[HTML]{FFFFFF} 
\cellcolor[HTML]{FFFFFF}                                                                   & \multirow{-2}{*}{\cellcolor[HTML]{FFFFFF}NOT IRONY} & \multirow{-2}{*}{\cellcolor[HTML]{FFFFFF}\textbf{0.92}} & \multirow{-2}{*}{\cellcolor[HTML]{FFFFFF}0.93}          & \multirow{-2}{*}{\cellcolor[HTML]{FFFFFF}0.93}          & \cellcolor[HTML]{FFFFFF}                       \\ \cline{2-5}
\rowcolor[HTML]{FFFFFF} 
\multirow{-5}{*}{\cellcolor[HTML]{FFFFFF}\textbf{dccuchile/bert-base-spanish-wwm-cased}}   & AVG                                                 & 0.82                                                    & \textbf{0.81}                                           & \textbf{0.82}                                           & \multirow{-5}{*}{\cellcolor[HTML]{FFFFFF}0.88} \\ \hline
\rowcolor[HTML]{FFFFFF} 
\cellcolor[HTML]{FFFFFF}                                                                   & \cellcolor[HTML]{FFFFFF}                            & \cellcolor[HTML]{FFFFFF}                                & \cellcolor[HTML]{FFFFFF}                                & \cellcolor[HTML]{FFFFFF}                                & \cellcolor[HTML]{FFFFFF}                       \\
\rowcolor[HTML]{FFFFFF} 
\cellcolor[HTML]{FFFFFF}                                                                   & \multirow{-2}{*}{\cellcolor[HTML]{FFFFFF}IRONY}     & \multirow{-2}{*}{\cellcolor[HTML]{FFFFFF}0.68}          & \multirow{-2}{*}{\cellcolor[HTML]{FFFFFF}\textbf{0.67}} & \multirow{-2}{*}{\cellcolor[HTML]{FFFFFF}\textbf{0.67}} & \cellcolor[HTML]{FFFFFF}                       \\ \cline{2-5}
\rowcolor[HTML]{FFFFFF} 
\cellcolor[HTML]{FFFFFF}                                                                   & \cellcolor[HTML]{FFFFFF}                            & \cellcolor[HTML]{FFFFFF}                                & \cellcolor[HTML]{FFFFFF}                                & \cellcolor[HTML]{FFFFFF}                                & \cellcolor[HTML]{FFFFFF}                       \\
\rowcolor[HTML]{FFFFFF} 
\cellcolor[HTML]{FFFFFF}                                                                   & \multirow{-2}{*}{\cellcolor[HTML]{FFFFFF}NOT IRONY} & \multirow{-2}{*}{\cellcolor[HTML]{FFFFFF}\textbf{0.91}} & \multirow{-2}{*}{\cellcolor[HTML]{FFFFFF}0.92}          & \multirow{-2}{*}{\cellcolor[HTML]{FFFFFF}0.91}          & \cellcolor[HTML]{FFFFFF}                       \\ \cline{2-5}
\rowcolor[HTML]{FFFFFF} 
\multirow{-5}{*}{\cellcolor[HTML]{FFFFFF}\textbf{beto-cased-finetuned-xix-latam}}          & AVG                                                 & 0.80                                           & \textbf{0.79}                                           & \textbf{0.79}                                           & \multirow{-5}{*}{\cellcolor[HTML]{FFFFFF}0.87} \\ \hline
\end{tabular}%
}
\caption{\textit{Results of the BERT-based Classification Pipeline on the \textit{AUGMENTED} dataset for the binary task.}}
\label{tab:my-table8}
\end{table}

In the binary classification approach, the results in Table \ref{tab:my-table8} showed substantial improvement in the typical precision increase compared to multi-class experiments and in addressing the usual problem of low recall. The model dccuchile/bert-base-spanish-wwm-cased demonstrates an F1 score for the 'IRONY' class above any other obtained in the experiments.

\section{Discussion}
 
As expected, the baseline using GPT-4o exclusively for classification demonstrated the weakest performance among the models. Its accuracy and recall metrics were significantly lower than those based on our BERT-based pipeline and data augmentation techniques, with accuracy values of 0.39 for multi-class and 0.72 for binary classification. This was particularly evident in multi-class classification, where irony detection proved especially challenging. These results highlight the limitations of relying solely on GPT-4o for nuanced text analysis tasks.

The binary classification approach delivered superior overall performance, with accuracy values approaching one hundred percent. Simplifying the task to binary classification (ironic vs. non-ironic) improved generalization and precision across most configurations. However, the setup consistently struggled with recall for the irony class, which ranged from 0.09 to a maximum of 0.34 in baseline configurations. This indicates that while the model accurately predicted non-ironic cases, it frequently failed to recognize ironic instances.

The results of the semi-automated annotation process emphasize the valuable role of human verification in supporting automated annotation methods. The evaluation through human inspection of GPT-4o suggestions for tagging irony revealed some differences between machine-generated tags and human evaluations, as shown in Table \ref{tab:human-gpt-comparison}. While GPT-4o marked 73.6\% of entries as ironic, human evaluators assigned this tag to only 53.1\% of entries. Additionally, there was a notable difference in the tagging of negative, positive, and neutral sentiments, with humans detecting more negative sentiments and some positive sentiments that GPT-4o overlooked. GPT-4o struggled with some entries with low-quality OCR transcriptions, misleading the model to produce hallucinations. These entries were introduced in the 'unreadable' category—1.7\% and were excluded from the final dataset.

\begin{table}[H]
\centering
\resizebox{0.28\textwidth}{!}{%
\begin{tabular}{|l|c|c|}
\hline
\multicolumn{1}{|c|}{\textit{\textbf{Tag}}} & \textit{\textbf{GPT-4o Tag}} & \textit{\textbf{Human Tag}} \\ \hline
\textbf{Irony}                              & 73.6\%                    & 53.1\%                      \\ \hline
\textbf{Negative}                           & 3.9\%                     & 13.1\%                      \\ \hline
\textbf{Positive}                           & 0\%                       & 2.9\%                       \\ \hline
\textbf{Neutral}                            & 22.6\%                    & 29.1\%                      \\ \hline
\textbf{Unreadable}                         & -                         & 1.7\%                       \\ \hline
\end{tabular}%
}
\caption{\textit{Comparison of GPT-4o and Human Tags.}}
\label{tab:human-gpt-comparison}
\end{table}

The disparity in positive sentiment detection highlights GPT-4o's limitation in contextualizing historical nuances. For instance, the model struggled to discern an author's potentially positive intent mostly because, during that period, poetic language was often employed to praise individuals or concepts, which was not indicative of irony. However, in modern contexts, such excessive praise, particularly concerning politics, might typically be interpreted as ironic, highlighting the differing interpretations across eras. These findings further illustrate GPT-4o's cultural and historical biases, suggesting that while these models provide substantial assistance, human expertise remains essential for accurate sentiment analysis in complex historical datasets.

The process facilitated the generation of classifications by processing a set of previously untagged entries with a higher likelihood of irony using a tailored GPT-4o prompt. The GPT-4o automatic classifications and the detailed justifications were reviewed by human experts for accuracy. Confirmed annotations were integrated into the PRIMARY dataset, enriching and balancing it. This effectively addressed class imbalance as shown in Table \ref{tab:my-tableBalance}, a problematic limitation in earlier experiments, while expanding the dataset with additional examples of ironic content.

\begin{table}[H]
\centering
\resizebox{0.28\textwidth}{!}{%
\begin{tabular}{|l|c|c|}
\hline
\multicolumn{1}{|c|}{\textit{\textbf{Category}}} & \textit{\textbf{Primary}} & \textit{\textbf{Augmented}} \\ \hline
\textbf{Irony}                                   & 10.68\%                   & 22.40\%                     \\ \hline
\textbf{Negative}                                & 25.64\%                   & 22.32\%                     \\ \hline
\textbf{Neutral}                                 & 28.89\%                   & 29.12\%                     \\ \hline
\textbf{Positive}                                & 34.78\%                   & 26.16\%                     \\ \hline
\end{tabular}%
}
\caption{\textit{Classes Distribution in the PRIMARY and the AUGMENTED datasets.}}
\label{tab:my-tableBalance}
\end{table}

For multi-class classification, the \textit{AUGMENTED} dataset led to higher recall values for irony detection, suggesting that the semi-automated annotation strategy contributed positively. However, the performance of non-ironic classes showed slight reductions, highlighting the trade-offs involved in emphasizing irony-related signals. In binary classification, combining the semi-automated annotation process with the dccuchile/bert-base-spanish-wwm-cased model yielded strong results, with precision and recall values of 0.70 and 0.93, respectively. Notably, recall improved significantly, increasing by up to 0.50 compared to previous experiments. These results indicate that the semi-automated process helped expand the dataset and mitigate imbalance, enhancing irony detection\footnote{The model is available at \url{https://huggingface.co/Flaglab/latam-xix-irony} Detailed processing steps can be found at \url{https://github.com/historicalink/ironydetection}}.

\begin{figure}[t]
\centering
  \includegraphics[width=0.8\linewidth]{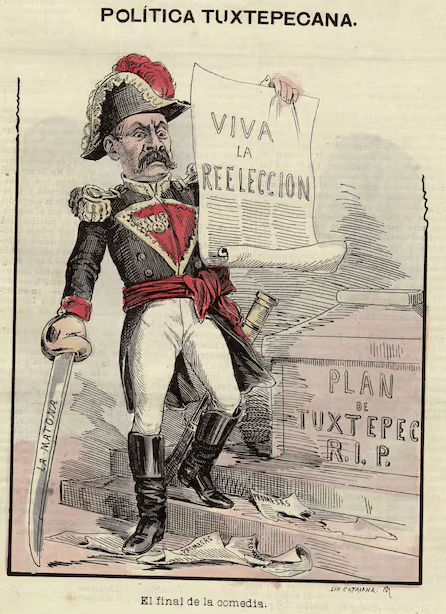}
  \caption{An example of a cartoon depicting the ironic situation lived during the time of \textit{porfiriato} in Mexico. El Hijo del Ahuizote, Mexico. Dec 7th, 1890. From the cartoons exhibition at Museo del Estanquillo in 2024. }
  \label{fig:porfirio}
\end{figure}

Interestingly, the \textit{ENHANCED} dataset, designed for emotional and contextual enrichment, while effectively sharpening certain sentiment cues, did not result in significant gains in irony detection over the original dataset. This finding indicates that enhancing emotional intensity alone does not necessarily capture the subtleties of historical irony. For example, GPT-4o struggled to identify irony when applying enhancement techniques to a sentence expressing political contradictions during the \textit{Porfiriato} in Mexico\footnote{ The translated sentence read: \textit{'Come, we said with enthusiasm, he is the one who will put us on the horns of the moon, with his respect for the law, his pure patriotism, and his famed honesty. You will see what Mr. Porfirio can do with the Tuxtepec Plan. We were already quite satisfied with our man in power, and everything was set and of good quality.'}
The original text in Spanish is: '\textit{Ven, dijimos con entusiasmo, es el que nos va a poner en los cuernos de la luna, con su respeto a la ley, y su puro patriotismo y su mentada honradez. Verán lo que es D. Porfirio con el plan de Tuxtepecl. Estábamos ya muy anchos con nuestro hombre en el poder y ya con toda la cosa muy lista y de buen jaéz}'}, primarily due to a lack of historical context. As depicted in Figure \ref{fig:porfirio}, Porfirio Díaz, who initially opposed re-elections citing constitutional violations, ironically remained in power for 35 years. Although enhancing has been successful in previous works \cite{Lin2023}, it did not effectively capture the specific cultural and historical features present in Latin-American historical texts. This highlights the cultural bias embedded in commercial models such as GPT-4o, emphasizing the need for more tailored approaches when working with historical and culturally nuanced materials.

Overall, the semi-automated annotation process, especially in binary classification, achieved the best performance for irony detection. This approach's ability to expand the dataset and address class imbalance marks the best approach we found for this task.

Despite these advances, irony detection continues to present significant challenges due to its inherent complexity. Historical texts bring additional layers of difficulty with their unique linguistic and cultural references. Future research should focus on refining domain-specific prompts, evaluating alternative models, and developing increasingly automated architectures, including agent-based workflows that systematically incorporate historical context to enhance irony detection accuracy. Continued efforts to expand and enrich historical datasets will contribute to more reliable and generalized methods for irony detection and sentiment analysis across diverse cultural and historical contexts, ultimately enriching both humanities scholarship and the capabilities of large LLMs.

\section{Acknowledgements}

We thank the two anonymous NAACL 2025 NLP4DH conference reviewers for their helpful feedback and suggestions.
\section{Limitations}

Although five different models were employed in the classification experiments, a notable limitation remains the reliance on GPT-4o for the initial dataset enhancement and augmentation steps. While this reliance does not compromise the significance and robustness of our findings, it is costly and restricts scalability. Future research could benefit from exploring alternative models and systematically comparing their effectiveness in semi-automated data augmentation tasks, aiming to identify options that are both accessible and cost-effective for broader implementation.

\bibliography{references} 

\begin{appendices}
\section{Prompt used to enhance the sentiment and context of the dataset} \label{prompt1}
"\textit{Expande este texto de manera de que mantenga su significado original, se debe hacer mucho énfasis en la carga emocional del texto, de manera que la versión final obtenida permita una mejor identificación del sentimiento general del mismo. Únicamente responde con el texto expandido y esfuérzate por conservar la sintaxis y morfología del español latinoamericano del siglo 19}"

\subsection{Input and Output Examples}
\label{propmtex1}
\textbf{Input text:} 
"\textit{habla de noventa y nueve anos, y noventa y nueve mas de transcurso para que el gobierno pueda entrar en posesion de el, es decir, dos siglos, en cuyo tiempo, probablemente, cuanto a la tarifa de trasportes, q discrecion del sr.  cisneros establecerla durante el tiempo de la construccion del camino, y diez anos mas y mientras que antes se obligaba a cobrar solamente la tercera parte de la tarifa por las tropas del gobierno y los efectos del servicio publico, posteriornente exige la mitad de ella.  una tarifa arbitraria y no establecida con previo acuerdo para ejercer toda la extorsion que se quiera.  aun faltan por analizar algunos otros puntos importantes  pero por temor de cansa vuestra atencion, y por haber sido ya, repetiperiodica, prescindimos de ocuparnos de ellos.  es 6 todas luces de vital importancia la continuacion del ferrocarril de girardot mas, en nuestro humilde concepto, es preferi.  ble a un contrato sumamente gravoso, que la verifique el gobierno por administracion  pues que a la vista esta la mejora de la parte de la direccion, y, porque de ello es tambien miguel guerrero y la acertada direccion del inteligente ingeniero d.}"

\textbf{Output text:} "\textit{Habla de noventa y nueve años, y noventa y nueve más de transcurso para que el gobierno pueda entrar en posesión de él, es decir, dos siglos completos, en cuyo tiempo, probablemente, en cuanto a la tarifa de transportes, queda a discreción del señor Cisneros establecerla durante el tiempo de la construcción del camino, y diez años más. Y mientras que antes se obligaba a cobrar solamente la tercera parte de la tarifa por las tropas del gobierno y los efectos del servicio público, posteriormente se exige la mitad de ella. Una tarifa arbitraria y no establecida con previo acuerdo para ejercer toda la extorsión que se quiera. Aún faltan por analizar algunos otros puntos importantes, pero por temor de cansar vuestra atención, y por haber sido ya repetidos periódicamente, prescindimos de ocuparnos de ellos. Es a todas luces de vital importancia la continuación del ferrocarril de Girardot, mas, en nuestro humilde concepto, es preferible a un contrato sumamente gravoso, que la verifique el gobierno por administración, pues que a la vista está¡ la mejora de la parte de la dirección, y, porque de ello es también Miguel Guerrero y la acertada dirección del inteligente ingeniero don.}"

\section{Prompt used for the semi-automatic labeling process (Translated from Spanish)} \footnote{Prompts in Spanish can be found at \url{https://github.com/historicalink/ironydetection}.} \label{ex1}

"\textit{A text in Spanish from 19th-century Latin American press will be received.}
\textit{This text may or may not contain some form of irony, meaning it fulfills one of the following situations:}
\textit{- It presents a contradiction between the reality described in the context and what is said.}
\textit{- It presents a contradiction between the historical reality of 19th-century Latin America and what is said.}
\textit{- It presents a contradiction between what is said and the tone in which it is said (based on the use of capitalization and punctuation).}

\textit{This text may contain a critique of a contradictory political or social situation that occurred, but it is not necessarily ironic; it could be a negative political opinion.
The text may also contain contradictory comparisons or hyperboles, but it is not necessarily ironic; it could be an expression with poetic language. For it to be an ironic contradiction, there must be an intent of humor or mockery in the text, not merely an intent of political critique or contradiction or an intent of figurative or poetic description.}

\textit{The task is to identify whether there is irony present in any contradiction in the text and explain why it is contradictory and what the author’s intention is. If no irony is detected, you must explain why it is not irony and indicate whether the text has a positive, negative, or neutral sentiment.}
\textit{The response must begin with one of these 4 words based on the inference: “IRONY,” “POSITIVE,” “NEGATIVE,” “NEUTRAL,” written in single quotation marks (''). Next, the explanation of what the contradiction is (if ironic) or why it is not irony must be added between asterisks (*).}
\textit{You must not include anything beyond what is requested. The final response must not exceed 500 words, including the description."}

\subsection{Input and Output Examples } \label{propmtex}
\textbf{Input Text:} The same as in the appendix 
\ref{prompt1}.

\textbf{Output text:}
"\textit{'NEGATIVE' *The text provides a negative critique of the government's management and the arbitrary transportation fee imposed, but there is no comedic or mocking intention indicating irony. The contradiction mentioned constitutes a direct and serious criticism of the political and administrative situation, without elements of irony.*}"

\section{Tables}\label{ex4}
\subsection{Baselines Results}\label{ex5}
\begin{table}[H]
\resizebox{\columnwidth}{!}{%
\begin{tabular}{|c|l|c|c|c|c|}
\hline
\textit{\textbf{Model}}                                                                    & \multicolumn{1}{c|}{\textit{\textbf{Category}}} & \textit{\textbf{Precision}} & \textit{\textbf{Recall}} & \textit{\textbf{F1 Score}} & \textit{\textbf{Accuracy}}                     \\ \hline
\rowcolor[HTML]{FFFFFF} 
\cellcolor[HTML]{FFFFFF}                                                                   & IRONY                                           & 0.28                        & 0.11                     & 0.15                       & \cellcolor[HTML]{FFFFFF}                       \\ \cline{2-5}
\rowcolor[HTML]{FFFFFF} 
\cellcolor[HTML]{FFFFFF}                                                                   & NEGATIVE                                        & 0.47                        & 0.60                     & 0.53                       & \cellcolor[HTML]{FFFFFF}                       \\ \cline{2-5}
\rowcolor[HTML]{FFFFFF} 
\cellcolor[HTML]{FFFFFF}                                                                   & NEUTRAL                                         & 0.60                        & 0.54                     & 0.57                       & \cellcolor[HTML]{FFFFFF}                       \\ \cline{2-5}
\rowcolor[HTML]{FFFFFF} 
\cellcolor[HTML]{FFFFFF}                                                                   & POSITIVE                                        & 0.58                        & 0.63                     & 0.60                       & \cellcolor[HTML]{FFFFFF}                       \\ \cline{2-5}
\rowcolor[HTML]{FFFFFF} 
\multirow{-5}{*}{\cellcolor[HTML]{FFFFFF}\textbf{bert-base-uncased}}                       & W. AVG                                          & 0.52                        & 0.54                     & 0.52                       & \multirow{-5}{*}{\cellcolor[HTML]{FFFFFF}0.54} \\ \hline
\rowcolor[HTML]{FFFFFF} 
\cellcolor[HTML]{FFFFFF}                                                                   & IRONY                                           & 0.52                        & 0.34                     & 0.41                       & \cellcolor[HTML]{FFFFFF}                       \\ \cline{2-5}
\rowcolor[HTML]{FFFFFF} 
\cellcolor[HTML]{FFFFFF}                                                                   & NEGATIVE                                        & 0.49                        & 0.52                     & 0.50                       & \cellcolor[HTML]{FFFFFF}                       \\ \cline{2-5}
\rowcolor[HTML]{FFFFFF} 
\cellcolor[HTML]{FFFFFF}                                                                   & NEUTRAL                                         & 0.68                        & 0.60                     & 0.64                       & \cellcolor[HTML]{FFFFFF}                       \\ \cline{2-5}
\rowcolor[HTML]{FFFFFF} 
\cellcolor[HTML]{FFFFFF}                                                                   & POSITIVE                                        & 0.61                        & 0.70                     & 0.65                       & \cellcolor[HTML]{FFFFFF}                       \\ \cline{2-5}
\rowcolor[HTML]{FFFFFF} 
\multirow{-5}{*}{\cellcolor[HTML]{FFFFFF}\textbf{bert-base-multilingual-uncased}}          & W. AVG                                          & 0.59                        & 0.59                     & 0.58                       & \multirow{-5}{*}{\cellcolor[HTML]{FFFFFF}0.59} \\ \hline
\rowcolor[HTML]{FFFFFF} 
\cellcolor[HTML]{FFFFFF}                                                                   & \cellcolor[HTML]{FFFFFF}IRONY                   & 0.54                        & 0.43                     & 0.48                       & \cellcolor[HTML]{FFFFFF}                       \\ \cline{2-5}
\rowcolor[HTML]{FFFFFF} 
\cellcolor[HTML]{FFFFFF}                                                                   & \cellcolor[HTML]{FFFFFF}NEGATIVE                & 0.56                        & 0.59                     & 0.58                       & \cellcolor[HTML]{FFFFFF}                       \\ \cline{2-5}
\rowcolor[HTML]{FFFFFF} 
\cellcolor[HTML]{FFFFFF}                                                                   & \cellcolor[HTML]{FFFFFF}NEUTRAL                 & 0.74                        & 0.63                     & 0.68                       & \cellcolor[HTML]{FFFFFF}                       \\ \cline{2-5}
\rowcolor[HTML]{FFFFFF} 
\cellcolor[HTML]{FFFFFF}                                                                   & \cellcolor[HTML]{FFFFFF}POSITIVE                & 0.65                        & 0.76                     & 0.70                       & \cellcolor[HTML]{FFFFFF}                       \\ \cline{2-5}
\rowcolor[HTML]{FFFFFF} 
\multirow{-5}{*}{\cellcolor[HTML]{FFFFFF}\textbf{dccuchile/bert-base-spanish-wwm-uncased}} & \cellcolor[HTML]{FFFFFF}W. AVG                  & 0.64                        & 0.64                     & 0.64                       & \multirow{-5}{*}{\cellcolor[HTML]{FFFFFF}0.64} \\ \hline
\rowcolor[HTML]{FFFFFF} 
\cellcolor[HTML]{FFFFFF}                                                                   & IRONY                                           & 0.61                        & 0.47                     & 0.53                       & \cellcolor[HTML]{FFFFFF}                       \\ \cline{2-5}
\rowcolor[HTML]{FFFFFF} 
\cellcolor[HTML]{FFFFFF}                                                                   & NEGATIVE                                        & 0.60                        & 0.62                     & 0.61                       & \cellcolor[HTML]{FFFFFF}                       \\ \cline{2-5}
\rowcolor[HTML]{FFFFFF} 
\cellcolor[HTML]{FFFFFF}                                                                   & NEUTRAL                                         & 0.72                        & 0.66                     & 0.69                       & \cellcolor[HTML]{FFFFFF}                       \\ \cline{2-5}
\rowcolor[HTML]{FFFFFF} 
\cellcolor[HTML]{FFFFFF}                                                                   & POSITIVE                                        & 0.66                        & 0.75                     & 0.70                       & \cellcolor[HTML]{FFFFFF}                       \\ \cline{2-5}
\rowcolor[HTML]{FFFFFF} 
\multirow{-5}{*}{\cellcolor[HTML]{FFFFFF}\textbf{dccuchile/bert-base-spanish-wwm-cased}}   & W. AVG                                          & 0.66                        & 0.66                     & 0.65                       & \multirow{-5}{*}{\cellcolor[HTML]{FFFFFF}0.66} \\ \hline
\rowcolor[HTML]{FFFFFF} 
\cellcolor[HTML]{FFFFFF}                                                                   & \cellcolor[HTML]{FFFFFF}IRONY                   & 0.59                        & 0.47                     & 0.52                       & \cellcolor[HTML]{FFFFFF}                       \\ \cline{2-5}
\rowcolor[HTML]{FFFFFF} 
\cellcolor[HTML]{FFFFFF}                                                                   & \cellcolor[HTML]{FFFFFF}NEGATIVE                & 0.56                        & 0.60                     & 0.58                       & \cellcolor[HTML]{FFFFFF}                       \\ \cline{2-5}
\rowcolor[HTML]{FFFFFF} 
\cellcolor[HTML]{FFFFFF}                                                                   & \cellcolor[HTML]{FFFFFF}NEUTRAL                 & 0.71                        & 0.59                     & 0.64                       & \cellcolor[HTML]{FFFFFF}                       \\ \cline{2-5}
\rowcolor[HTML]{FFFFFF} 
\cellcolor[HTML]{FFFFFF}                                                                   & \cellcolor[HTML]{FFFFFF}POSITIVE                & 0.62                        & 0.74                     & 0.68                       & \cellcolor[HTML]{FFFFFF}                       \\ \cline{2-5}
\rowcolor[HTML]{FFFFFF} 
\multirow{-5}{*}{\cellcolor[HTML]{FFFFFF}\textbf{beto-cased-finetuned-xix-latam}}                      & \cellcolor[HTML]{FFFFFF}W. AVG                  & 0.63                        & 0.63                     & 0.63                       & \multirow{-5}{*}{\cellcolor[HTML]{FFFFFF}0.63} \\ \hline
\end{tabular}%
}
\caption{\textit{Baseline Results for multi-class classification}}
\label{tab:my-table9}
\end{table}

\begin{table}[H]
\resizebox{\columnwidth}{!}{%
\begin{tabular}{|c|l|c|c|c|c|}
\hline
\textit{\textbf{Model}}                                                                    & \multicolumn{1}{c|}{\textit{\textbf{Category}}}     & \textit{\textbf{Precision}}                    & \textit{\textbf{Recall}}                       & \textit{\textbf{F1 Score}}                     & \textit{\textbf{Accuracy}}                     \\ \hline
\rowcolor[HTML]{FFFFFF} 
\cellcolor[HTML]{FFFFFF}                                                                   & \cellcolor[HTML]{FFFFFF}                            & \cellcolor[HTML]{FFFFFF}                       & \cellcolor[HTML]{FFFFFF}                       & \cellcolor[HTML]{FFFFFF}                       & \cellcolor[HTML]{FFFFFF}                       \\
\rowcolor[HTML]{FFFFFF} 
\cellcolor[HTML]{FFFFFF}                                                                   & \multirow{-2}{*}{\cellcolor[HTML]{FFFFFF}IRONY}     & \multirow{-2}{*}{\cellcolor[HTML]{FFFFFF}0.80} & \multirow{-2}{*}{\cellcolor[HTML]{FFFFFF}0.09} & \multirow{-2}{*}{\cellcolor[HTML]{FFFFFF}0.15} & \cellcolor[HTML]{FFFFFF}                       \\ \cline{2-5}
\rowcolor[HTML]{FFFFFF} 
\cellcolor[HTML]{FFFFFF}                                                                   & \cellcolor[HTML]{FFFFFF}                            & \cellcolor[HTML]{FFFFFF}                       & \cellcolor[HTML]{FFFFFF}                       & \cellcolor[HTML]{FFFFFF}                       & \cellcolor[HTML]{FFFFFF}                       \\
\rowcolor[HTML]{FFFFFF} 
\cellcolor[HTML]{FFFFFF}                                                                   & \multirow{-2}{*}{\cellcolor[HTML]{FFFFFF}NOT IRONY} & \multirow{-2}{*}{\cellcolor[HTML]{FFFFFF}0.89} & \multirow{-2}{*}{\cellcolor[HTML]{FFFFFF}1.00} & \multirow{-2}{*}{\cellcolor[HTML]{FFFFFF}0.94} & \cellcolor[HTML]{FFFFFF}                       \\ \cline{2-5}
\rowcolor[HTML]{FFFFFF} 
\multirow{-5}{*}{\cellcolor[HTML]{FFFFFF}\textbf{bert-base-uncased}}                       & AVG                                                 & 0.85                                           & 0.54                                           & 0.55                                           & \multirow{-5}{*}{\cellcolor[HTML]{FFFFFF}0.89} \\ \hline
\rowcolor[HTML]{FFFFFF} 
\cellcolor[HTML]{FFFFFF}                                                                   & \cellcolor[HTML]{FFFFFF}                            & \cellcolor[HTML]{FFFFFF}                       & \cellcolor[HTML]{FFFFFF}                       & \cellcolor[HTML]{FFFFFF}                       & \cellcolor[HTML]{FFFFFF}                       \\
\rowcolor[HTML]{FFFFFF} 
\cellcolor[HTML]{FFFFFF}                                                                   & \multirow{-2}{*}{\cellcolor[HTML]{FFFFFF}IRONY}     & \multirow{-2}{*}{\cellcolor[HTML]{FFFFFF}0.75} & \multirow{-2}{*}{\cellcolor[HTML]{FFFFFF}0.32} & \multirow{-2}{*}{\cellcolor[HTML]{FFFFFF}0.45} & \cellcolor[HTML]{FFFFFF}                       \\ \cline{2-5}
\rowcolor[HTML]{FFFFFF} 
\cellcolor[HTML]{FFFFFF}                                                                   & \cellcolor[HTML]{FFFFFF}                            & \cellcolor[HTML]{FFFFFF}                       & \cellcolor[HTML]{FFFFFF}                       & \cellcolor[HTML]{FFFFFF}                       & \cellcolor[HTML]{FFFFFF}                       \\
\rowcolor[HTML]{FFFFFF} 
\cellcolor[HTML]{FFFFFF}                                                                   & \multirow{-2}{*}{\cellcolor[HTML]{FFFFFF}NOT IRONY} & \multirow{-2}{*}{\cellcolor[HTML]{FFFFFF}0.92} & \multirow{-2}{*}{\cellcolor[HTML]{FFFFFF}0.99} & \multirow{-2}{*}{\cellcolor[HTML]{FFFFFF}0.95} & \cellcolor[HTML]{FFFFFF}                       \\ \cline{2-5}
\rowcolor[HTML]{FFFFFF} 
\multirow{-5}{*}{\cellcolor[HTML]{FFFFFF}\textbf{bert-base-multilingual-uncased}}          & AVG                                                 & 0.83                                           & 0.65                                           & 0.70                                           & \multirow{-5}{*}{\cellcolor[HTML]{FFFFFF}0.91} \\ \hline
\rowcolor[HTML]{FFFFFF} 
\cellcolor[HTML]{FFFFFF}                                                                   & \cellcolor[HTML]{FFFFFF}                            & \cellcolor[HTML]{FFFFFF}                       & \cellcolor[HTML]{FFFFFF}                       & \cellcolor[HTML]{FFFFFF}                       & \cellcolor[HTML]{FFFFFF}                       \\
\rowcolor[HTML]{FFFFFF} 
\cellcolor[HTML]{FFFFFF}                                                                   & \multirow{-2}{*}{\cellcolor[HTML]{FFFFFF}IRONY}     & \multirow{-2}{*}{\cellcolor[HTML]{FFFFFF}0.79} & \multirow{-2}{*}{\cellcolor[HTML]{FFFFFF}0.23} & \multirow{-2}{*}{\cellcolor[HTML]{FFFFFF}0.36} & \cellcolor[HTML]{FFFFFF}                       \\ \cline{2-5}
\rowcolor[HTML]{FFFFFF} 
\cellcolor[HTML]{FFFFFF}                                                                   & \cellcolor[HTML]{FFFFFF}                            & \cellcolor[HTML]{FFFFFF}                       & \cellcolor[HTML]{FFFFFF}                       & \cellcolor[HTML]{FFFFFF}                       & \cellcolor[HTML]{FFFFFF}                       \\
\rowcolor[HTML]{FFFFFF} 
\cellcolor[HTML]{FFFFFF}                                                                   & \multirow{-2}{*}{\cellcolor[HTML]{FFFFFF}NOT IRONY} & \multirow{-2}{*}{\cellcolor[HTML]{FFFFFF}0.91} & \multirow{-2}{*}{\cellcolor[HTML]{FFFFFF}0.99} & \multirow{-2}{*}{\cellcolor[HTML]{FFFFFF}0.95} & \cellcolor[HTML]{FFFFFF}                       \\ \cline{2-5}
\rowcolor[HTML]{FFFFFF} 
\multirow{-5}{*}{\cellcolor[HTML]{FFFFFF}\textbf{dccuchile/bert-base-spanish-wwm-uncased}} & GENERAL                                             & 0.85                                           & 0.61                                           & 0.65                                           & \multirow{-5}{*}{\cellcolor[HTML]{FFFFFF}0.91} \\ \hline
\rowcolor[HTML]{FFFFFF} 
\cellcolor[HTML]{FFFFFF}                                                                   & \cellcolor[HTML]{FFFFFF}                            & \cellcolor[HTML]{FFFFFF}                       & \cellcolor[HTML]{FFFFFF}                       & \cellcolor[HTML]{FFFFFF}                       & \cellcolor[HTML]{FFFFFF}                       \\
\rowcolor[HTML]{FFFFFF} 
\cellcolor[HTML]{FFFFFF}                                                                   & \multirow{-2}{*}{\cellcolor[HTML]{FFFFFF}IRONY}     & \multirow{-2}{*}{\cellcolor[HTML]{FFFFFF}0.80} & \multirow{-2}{*}{\cellcolor[HTML]{FFFFFF}0.34} & \multirow{-2}{*}{\cellcolor[HTML]{FFFFFF}0.48} & \cellcolor[HTML]{FFFFFF}                       \\ \cline{2-5}
\rowcolor[HTML]{FFFFFF} 
\cellcolor[HTML]{FFFFFF}                                                                   & \cellcolor[HTML]{FFFFFF}                            & \cellcolor[HTML]{FFFFFF}                       & \cellcolor[HTML]{FFFFFF}                       & \cellcolor[HTML]{FFFFFF}                       & \cellcolor[HTML]{FFFFFF}                       \\
\rowcolor[HTML]{FFFFFF} 
\cellcolor[HTML]{FFFFFF}                                                                   & \multirow{-2}{*}{\cellcolor[HTML]{FFFFFF}NOT IRONY} & \multirow{-2}{*}{\cellcolor[HTML]{FFFFFF}0.92} & \multirow{-2}{*}{\cellcolor[HTML]{FFFFFF}0.99} & \multirow{-2}{*}{\cellcolor[HTML]{FFFFFF}0.95} & \cellcolor[HTML]{FFFFFF}                       \\ \cline{2-5}
\rowcolor[HTML]{FFFFFF} 
\multirow{-5}{*}{\cellcolor[HTML]{FFFFFF}\textbf{dccuchile/bert-base-spanish-wwm-cased}}   & AVG                                                 & 0.86                                           & 0.66                                           & 0.72                                           & \multirow{-5}{*}{\cellcolor[HTML]{FFFFFF}0.91} \\ \hline
\rowcolor[HTML]{FFFFFF} 
\cellcolor[HTML]{FFFFFF}                                                                   & \cellcolor[HTML]{FFFFFF}                            & \cellcolor[HTML]{FFFFFF}                       & \cellcolor[HTML]{FFFFFF}                       & \cellcolor[HTML]{FFFFFF}                       & \cellcolor[HTML]{FFFFFF}                       \\
\rowcolor[HTML]{FFFFFF} 
\cellcolor[HTML]{FFFFFF}                                                                   & \multirow{-2}{*}{\cellcolor[HTML]{FFFFFF}IRONY}     & \multirow{-2}{*}{\cellcolor[HTML]{FFFFFF}0.71} & \multirow{-2}{*}{\cellcolor[HTML]{FFFFFF}0.21} & \multirow{-2}{*}{\cellcolor[HTML]{FFFFFF}0.33} & \cellcolor[HTML]{FFFFFF}                       \\ \cline{2-5}
\rowcolor[HTML]{FFFFFF} 
\cellcolor[HTML]{FFFFFF}                                                                   & \cellcolor[HTML]{FFFFFF}                            & \cellcolor[HTML]{FFFFFF}                       & \cellcolor[HTML]{FFFFFF}                       & \cellcolor[HTML]{FFFFFF}                       & \cellcolor[HTML]{FFFFFF}                       \\
\rowcolor[HTML]{FFFFFF} 
\cellcolor[HTML]{FFFFFF}                                                                   & \multirow{-2}{*}{\cellcolor[HTML]{FFFFFF}NOT IRONY} & \multirow{-2}{*}{\cellcolor[HTML]{FFFFFF}0.91} & \multirow{-2}{*}{\cellcolor[HTML]{FFFFFF}0.99} & \multirow{-2}{*}{\cellcolor[HTML]{FFFFFF}0.95} & \cellcolor[HTML]{FFFFFF}                       \\ \cline{2-5}
\rowcolor[HTML]{FFFFFF} 
\multirow{-5}{*}{\cellcolor[HTML]{FFFFFF}\textbf{beto-cased-finetuned-xix-latam}}                      & AVG                                                 & 0.81                                           & 0.60                                           & 0.64                                           & \multirow{-5}{*}{\cellcolor[HTML]{FFFFFF}0.90} \\ \hline
\end{tabular}
}
\caption{\textit{Baseline Results for binary classification}}
\label{tab:my-table10}
\end{table}
\subsection{Data Enhancement Results}\label{ex6}
\begin{table}[H]
\resizebox{\columnwidth}{!}{%
\begin{tabular}{|
>{\columncolor[HTML]{FFFFFF}}c |
>{\columncolor[HTML]{FFFFFF}}l |
>{\columncolor[HTML]{FFFFFF}}c |
>{\columncolor[HTML]{FFFFFF}}c |
>{\columncolor[HTML]{FFFFFF}}c |
>{\columncolor[HTML]{FFFFFF}}c |}
\hline
\textit{\textbf{Model}}                                                                    & \multicolumn{1}{c|}{\cellcolor[HTML]{FFFFFF}\textit{\textbf{Category}}} & \textit{Precision} & \textit{Recall} & \textit{F1 Score} & \textit{Accuracy}                              \\ \hline
\cellcolor[HTML]{FFFFFF}                                                                   & IRONY                                                                   & \textbf{0.65}      & \textbf{0.23}   & \textbf{0.34}     & \cellcolor[HTML]{FFFFFF}                       \\ \cline{2-5}
\cellcolor[HTML]{FFFFFF}                                                                   & NEGATIVE                                                                & \textbf{0.47}      & \textbf{0.64}   & \textbf{0.54}     & \cellcolor[HTML]{FFFFFF}                       \\ \cline{2-5}
\cellcolor[HTML]{FFFFFF}                                                                   & NEUTRAL                                                                 & \textbf{0.62}      & 0.50            & 0.55              & \cellcolor[HTML]{FFFFFF}                       \\ \cline{2-5}
\cellcolor[HTML]{FFFFFF}                                                                   & POSITIVE                                                                & 0.52               & 0.58            & 0.55              & \cellcolor[HTML]{FFFFFF}                       \\ \cline{2-5}
\multirow{-5}{*}{\cellcolor[HTML]{FFFFFF}\textbf{bert-base-uncased}}                       & W. AVG                                                                  & \textbf{0.55}      & 0.53            & 0.52              & \multirow{-5}{*}{\cellcolor[HTML]{FFFFFF}0.53} \\ \hline
\cellcolor[HTML]{FFFFFF}                                                                   & IRONY                                                                   & \textbf{0.67}      & 0.21            & 0.32              & \cellcolor[HTML]{FFFFFF}                       \\ \cline{2-5}
\cellcolor[HTML]{FFFFFF}                                                                   & NEGATIVE                                                                & \textbf{0.50}      & \textbf{0.60}   & \textbf{0.55}     & \cellcolor[HTML]{FFFFFF}                       \\ \cline{2-5}
\cellcolor[HTML]{FFFFFF}                                                                   & NEUTRAL                                                                 & 0.65               & 0.59            & 0.62              & \cellcolor[HTML]{FFFFFF}                       \\ \cline{2-5}
\cellcolor[HTML]{FFFFFF}                                                                   & POSITIVE                                                                & 0.55               & 0.65            & 0.60              & \cellcolor[HTML]{FFFFFF}                       \\ \cline{2-5}
\multirow{-5}{*}{\cellcolor[HTML]{FFFFFF}\textbf{bert-base-multilingual-uncased}}          & W. AVG                                                                  & 0.58               & 0.57            & 0.56              & \multirow{-5}{*}{\cellcolor[HTML]{FFFFFF}0.57} \\ \hline
\cellcolor[HTML]{FFFFFF}                                                                   & IRONY                                                                   & 0.54               & 0.40            & 0.46              & \cellcolor[HTML]{FFFFFF}                       \\ \cline{2-5}
\cellcolor[HTML]{FFFFFF}                                                                   & NEGATIVE                                                                & \textbf{0.59}      & \textbf{0.64}   & \textbf{0.61}     & \cellcolor[HTML]{FFFFFF}                       \\ \cline{2-5}
\cellcolor[HTML]{FFFFFF}                                                                   & NEUTRAL                                                                 & 0.72               & 0.59            & 0.65              & \cellcolor[HTML]{FFFFFF}                       \\ \cline{2-5}
\cellcolor[HTML]{FFFFFF}                                                                   & POSITIVE                                                                & 0.61               & 0.73            & 0.66              & \cellcolor[HTML]{FFFFFF}                       \\ \cline{2-5}
\multirow{-5}{*}{\cellcolor[HTML]{FFFFFF}\textbf{dccuchile/bert-base-spanish-wwm-uncased}} & W. AVG                                                                  & 0.63               & 0.63            & 0.63              & \multirow{-5}{*}{\cellcolor[HTML]{FFFFFF}0.63} \\ \hline
\cellcolor[HTML]{FFFFFF}                                                                   & IRONY                                                                   & 0.53               & 0.45            & 0.48              & \cellcolor[HTML]{FFFFFF}                       \\ \cline{2-5}
\cellcolor[HTML]{FFFFFF}                                                                   & NEGATIVE                                                                & \textbf{0.62}      & \textbf{0.66}   & \textbf{0.64}     & \cellcolor[HTML]{FFFFFF}                       \\ \cline{2-5}
\cellcolor[HTML]{FFFFFF}                                                                   & NEUTRAL                                                                 & 0.72               & 0.55            & 0.63              & \cellcolor[HTML]{FFFFFF}                       \\ \cline{2-5}
\cellcolor[HTML]{FFFFFF}                                                                   & POSITIVE                                                                & 0.62               & \textbf{0.76}   & 0.68              & \cellcolor[HTML]{FFFFFF}                       \\ \cline{2-5}
\multirow{-5}{*}{\cellcolor[HTML]{FFFFFF}\textbf{dccuchile/bert-base-spanish-wwm-cased}}   & W. AVG                                                                  & 0.62               & 0.61            & 0.61              & \multirow{-5}{*}{\cellcolor[HTML]{FFFFFF}0.64} \\ \hline
\cellcolor[HTML]{FFFFFF}                                                                   & IRONY                                                                   & \textbf{0.65}      & \textbf{0.51}   & \textbf{0.57}     & \cellcolor[HTML]{FFFFFF}                       \\ \cline{2-5}
\cellcolor[HTML]{FFFFFF}                                                                   & NEGATIVE                                                                & 0.54               & \textbf{0.62}   & \textbf{0.58}     & \cellcolor[HTML]{FFFFFF}                       \\ \cline{2-5}
\cellcolor[HTML]{FFFFFF}                                                                   & NEUTRAL                                                                 & 0.69               & 0.53            & 0.60              & \cellcolor[HTML]{FFFFFF}                       \\ \cline{2-5}
\cellcolor[HTML]{FFFFFF}                                                                   & POSITIVE                                                                & 0.58               & 0.69            & 0.63              & \cellcolor[HTML]{FFFFFF}                       \\ \cline{2-5}
\multirow{-5}{*}{\cellcolor[HTML]{FFFFFF}\textbf{beto-cased-finetuned-xix-latam}}                      & W. AVG                                                                  & 0.61               & 0.60            & 0.60              & \multirow{-5}{*}{\cellcolor[HTML]{FFFFFF}0.60} \\ \hline
\end{tabular}%
}
\caption{\textit{Enhancement Results. Multi-class classification tasks}}
\label{tab:my-table11}
\end{table}

\begin{table}[H]
\resizebox{\columnwidth}{!}{%
\begin{tabular}{|
>{\columncolor[HTML]{FFFFFF}}c |
>{\columncolor[HTML]{FFFFFF}}l |
>{\columncolor[HTML]{FFFFFF}}c |
>{\columncolor[HTML]{FFFFFF}}c |
>{\columncolor[HTML]{FFFFFF}}c |
>{\columncolor[HTML]{FFFFFF}}c |}
\hline
\textit{\textbf{Model}}                                                                    & \multicolumn{1}{c|}{\cellcolor[HTML]{FFFFFF}\textit{\textbf{Category}}} & \textit{\textbf{Precision}}                             & \textit{\textbf{Recall}}                                & \textit{\textbf{F1 Score}}                              & \textit{\textbf{Accuracy}}                              \\ \hline
\cellcolor[HTML]{FFFFFF}                                                                   & \cellcolor[HTML]{FFFFFF}                                                & \cellcolor[HTML]{FFFFFF}                                & \cellcolor[HTML]{FFFFFF}                                & \cellcolor[HTML]{FFFFFF}                                & \cellcolor[HTML]{FFFFFF}                                \\
\cellcolor[HTML]{FFFFFF}                                                                   & \multirow{-2}{*}{\cellcolor[HTML]{FFFFFF}IRONY}                         & \multirow{-2}{*}{\cellcolor[HTML]{FFFFFF}0.70}          & \multirow{-2}{*}{\cellcolor[HTML]{FFFFFF}\textbf{0.15}} & \multirow{-2}{*}{\cellcolor[HTML]{FFFFFF}\textbf{0.25}} & \cellcolor[HTML]{FFFFFF}                                \\ \cline{2-5}
\cellcolor[HTML]{FFFFFF}                                                                   & \cellcolor[HTML]{FFFFFF}                                                & \cellcolor[HTML]{FFFFFF}                                & \cellcolor[HTML]{FFFFFF}                                & \cellcolor[HTML]{FFFFFF}                                & \cellcolor[HTML]{FFFFFF}                                \\
\cellcolor[HTML]{FFFFFF}                                                                   & \multirow{-2}{*}{\cellcolor[HTML]{FFFFFF}NOT IRONY}                     & \multirow{-2}{*}{\cellcolor[HTML]{FFFFFF}\textbf{0.90}} & \multirow{-2}{*}{\cellcolor[HTML]{FFFFFF}0.99}          & \multirow{-2}{*}{\cellcolor[HTML]{FFFFFF}0.94}          & \cellcolor[HTML]{FFFFFF}                                \\ \cline{2-5}
\multirow{-5}{*}{\cellcolor[HTML]{FFFFFF}\textbf{bert-base-uncased}}                       & AVG                                                                     & 0.80                                                    & \textbf{0.57}                                           & \textbf{0.59}                                           & \multirow{-5}{*}{\cellcolor[HTML]{FFFFFF}\textbf{0.90}} \\ \hline
\cellcolor[HTML]{FFFFFF}                                                                   & \cellcolor[HTML]{FFFFFF}                                                & \cellcolor[HTML]{FFFFFF}                                & \cellcolor[HTML]{FFFFFF}                                & \cellcolor[HTML]{FFFFFF}                                & \cellcolor[HTML]{FFFFFF}                                \\
\cellcolor[HTML]{FFFFFF}                                                                   & \multirow{-2}{*}{\cellcolor[HTML]{FFFFFF}IRONY}                         & \multirow{-2}{*}{\cellcolor[HTML]{FFFFFF}0.62}          & \multirow{-2}{*}{\cellcolor[HTML]{FFFFFF}0.21}          & \multirow{-2}{*}{\cellcolor[HTML]{FFFFFF}0.32}          & \cellcolor[HTML]{FFFFFF}                                \\ \cline{2-5}
\cellcolor[HTML]{FFFFFF}                                                                   & \cellcolor[HTML]{FFFFFF}                                                & \cellcolor[HTML]{FFFFFF}                                & \cellcolor[HTML]{FFFFFF}                                & \cellcolor[HTML]{FFFFFF}                                & \cellcolor[HTML]{FFFFFF}                                \\
\cellcolor[HTML]{FFFFFF}                                                                   & \multirow{-2}{*}{\cellcolor[HTML]{FFFFFF}NOT IRONY}                     & \multirow{-2}{*}{\cellcolor[HTML]{FFFFFF}0.91}          & \multirow{-2}{*}{\cellcolor[HTML]{FFFFFF}0.98}          & \multirow{-2}{*}{\cellcolor[HTML]{FFFFFF}0.94}          & \cellcolor[HTML]{FFFFFF}                                \\ \cline{2-5}
\multirow{-5}{*}{\cellcolor[HTML]{FFFFFF}\textbf{bert-base-multilingual-uncased}}          & AVG                                                                     & 0.77                                                    & 0.60                                                    & 0.63                                                    & \multirow{-5}{*}{\cellcolor[HTML]{FFFFFF}0.90}          \\ \hline
\cellcolor[HTML]{FFFFFF}                                                                   & \cellcolor[HTML]{FFFFFF}                                                & \cellcolor[HTML]{FFFFFF}                                & \cellcolor[HTML]{FFFFFF}                                & \cellcolor[HTML]{FFFFFF}                                & \cellcolor[HTML]{FFFFFF}                                \\
\cellcolor[HTML]{FFFFFF}                                                                   & \multirow{-2}{*}{\cellcolor[HTML]{FFFFFF}IRONY}                         & \multirow{-2}{*}{\cellcolor[HTML]{FFFFFF}0.65}          & \multirow{-2}{*}{\cellcolor[HTML]{FFFFFF}\textbf{0.32}} & \multirow{-2}{*}{\cellcolor[HTML]{FFFFFF}\textbf{0.43}} & \cellcolor[HTML]{FFFFFF}                                \\ \cline{2-5}
\cellcolor[HTML]{FFFFFF}                                                                   & \cellcolor[HTML]{FFFFFF}                                                & \cellcolor[HTML]{FFFFFF}                                & \cellcolor[HTML]{FFFFFF}                                & \cellcolor[HTML]{FFFFFF}                                & \cellcolor[HTML]{FFFFFF}                                \\
\cellcolor[HTML]{FFFFFF}                                                                   & \multirow{-2}{*}{\cellcolor[HTML]{FFFFFF}NOT IRONY}                     & \multirow{-2}{*}{\cellcolor[HTML]{FFFFFF}0.92}          & \multirow{-2}{*}{\cellcolor[HTML]{FFFFFF}0.98}          & \multirow{-2}{*}{\cellcolor[HTML]{FFFFFF}0.95}          & \cellcolor[HTML]{FFFFFF}                                \\ \cline{2-5}
\multirow{-5}{*}{\cellcolor[HTML]{FFFFFF}\textbf{dccuchile/bert-base-spanish-wwm-uncased}} & AVG                                                                     & 0.78                                                    & \textbf{0.65}                                           & \textbf{0.69}                                           & \multirow{-5}{*}{\cellcolor[HTML]{FFFFFF}0.90}          \\ \hline
\cellcolor[HTML]{FFFFFF}                                                                   & \cellcolor[HTML]{FFFFFF}                                                & \cellcolor[HTML]{FFFFFF}                                & \cellcolor[HTML]{FFFFFF}                                & \cellcolor[HTML]{FFFFFF}                                & \cellcolor[HTML]{FFFFFF}                                \\
\cellcolor[HTML]{FFFFFF}                                                                   & \multirow{-2}{*}{\cellcolor[HTML]{FFFFFF}IRONY}                         & \multirow{-2}{*}{\cellcolor[HTML]{FFFFFF}0.58}          & \multirow{-2}{*}{\cellcolor[HTML]{FFFFFF}0.15}          & \multirow{-2}{*}{\cellcolor[HTML]{FFFFFF}0.24}          & \cellcolor[HTML]{FFFFFF}                                \\ \cline{2-5}
\cellcolor[HTML]{FFFFFF}                                                                   & \cellcolor[HTML]{FFFFFF}                                                & \cellcolor[HTML]{FFFFFF}                                & \cellcolor[HTML]{FFFFFF}                                & \cellcolor[HTML]{FFFFFF}                                & \cellcolor[HTML]{FFFFFF}                                \\
\cellcolor[HTML]{FFFFFF}                                                                   & \multirow{-2}{*}{\cellcolor[HTML]{FFFFFF}NOT IRONY}                     & \multirow{-2}{*}{\cellcolor[HTML]{FFFFFF}0.90}          & \multirow{-2}{*}{\cellcolor[HTML]{FFFFFF}0.99}          & \multirow{-2}{*}{\cellcolor[HTML]{FFFFFF}0.94}          & \cellcolor[HTML]{FFFFFF}                                \\ \cline{2-5}
\multirow{-5}{*}{\cellcolor[HTML]{FFFFFF}\textbf{dccuchile/bert-base-spanish-wwm-cased}}   & AVG                                                                     & 0.74                                                    & 0.57                                                    & 0.59                                                    & \multirow{-5}{*}{\cellcolor[HTML]{FFFFFF}0.89}          \\ \hline
\cellcolor[HTML]{FFFFFF}                                                                   & \cellcolor[HTML]{FFFFFF}                                                & \cellcolor[HTML]{FFFFFF}                                & \cellcolor[HTML]{FFFFFF}                                & \cellcolor[HTML]{FFFFFF}                                & \cellcolor[HTML]{FFFFFF}                                \\
\cellcolor[HTML]{FFFFFF}                                                                   & \multirow{-2}{*}{\cellcolor[HTML]{FFFFFF}IRONY}                         & \multirow{-2}{*}{\cellcolor[HTML]{FFFFFF}0.50}          & \multirow{-2}{*}{\cellcolor[HTML]{FFFFFF}0.17}          & \multirow{-2}{*}{\cellcolor[HTML]{FFFFFF}0.25}          & \cellcolor[HTML]{FFFFFF}                                \\ \cline{2-5}
\cellcolor[HTML]{FFFFFF}                                                                   & \cellcolor[HTML]{FFFFFF}                                                & \cellcolor[HTML]{FFFFFF}                                & \cellcolor[HTML]{FFFFFF}                                & \cellcolor[HTML]{FFFFFF}                                & \cellcolor[HTML]{FFFFFF}                                \\
\cellcolor[HTML]{FFFFFF}                                                                   & \multirow{-2}{*}{\cellcolor[HTML]{FFFFFF}NOT IRONY}                     & \multirow{-2}{*}{\cellcolor[HTML]{FFFFFF}0.90}          & \multirow{-2}{*}{\cellcolor[HTML]{FFFFFF}0.98}          & \multirow{-2}{*}{\cellcolor[HTML]{FFFFFF}0.94}          & \cellcolor[HTML]{FFFFFF}                                \\ \cline{2-5}
\multirow{-5}{*}{\cellcolor[HTML]{FFFFFF}\textbf{beto-cased-finetuned-xix-latam}}                      & AVG                                                                     & 0.70                                                    & 0.57                                                    & 0.60                                                    & \multirow{-5}{*}{\cellcolor[HTML]{FFFFFF}0.89}          \\ \hline
\end{tabular}%
}
\caption{\textit{Enhancement Results. Binary classification tasks}}
\label{tab:my-table12}
\end{table}

\end{appendices}
\end{document}